\documentclass[journal]{IEEEtran}
\ifCLASSINFOpdf
\else
\fi
\usepackage{multirow}
\usepackage{amssymb,amsfonts}
\usepackage{amsmath,graphicx}
\usepackage{caption}
\usepackage[]{hyperref}
\newcommand{\subparagraph}{}

\begin{document}
%
\title{Mask Combination of Multi-layer Graphs for Global Structure Inference}
%
%
%

\author{Eda~Bayram,~\IEEEmembership{}
\thanks{E. Bayram and P. Frossard are with the Signal Processing Laboratory (LTS4), EPFL, 1015 Lausanne, Switzerland
(e-mail: {eda.bayram, pascal.frossard}@epfl.ch).}
        Dorina~Thanou,~\IEEEmembership{}
\thanks{D. Thanou is with the Swiss Data Science Center, EPFL/ETHZ, 1015
Lausanne, Switzerland (e-mail: dorina.thanou@epfl.ch).}
        Elif~Vural,~\IEEEmembership{}
\thanks{E. Vural is with the Department of Electrical and
Electronics Engineering, METU, 06800 Ankara, Turkey (e-mail: velif@metu.edu.tr).}
        and~Pascal~Frossard~\IEEEmembership{}}

\maketitle

\begin{abstract}
Structure inference is an important task for network data processing and analysis in data science. In recent years, quite a few approaches have been developed to learn the graph structure underlying a set of observations captured in a data space.
Although real-world data is often acquired in settings where relationships are influenced by a priori known rules, such domain knowledge is still not well exploited in structure inference problems.
In this paper, we identify the structure of signals defined in a data space whose inner relationships are encoded by multi-layer graphs. We aim at properly exploiting the information originating from each layer to infer the global structure underlying the signals.
We thus present a novel method for combining the multiple graphs into a global graph using mask matrices, which are estimated through an optimization problem that accommodates the multi-layer graph information and a signal representation model.
The proposed mask combination method also estimates the contribution of each graph layer in the structure of signals.
The experiments conducted both on synthetic and real-world data suggest that integrating the multi-layer graph representation of the data in the structure inference framework enhances the learning procedure considerably by adapting to the quality and the quantity of the input data. 
\end{abstract}


%

\section{Introduction}
\IEEEPARstart{M}{any} real-world data can be represented with multiple forms of relations between data samples.
Examples include social networks that relate individuals based on different types of connections or behavioral similarities \cite{cozzo2016multilayer, wasserman1994social}, biological networks where different modes of interactions exist between neurons or brain regions \cite{bentley2016multilayer, de2017multilayer}, transportation networks which lead to the movement of people via different transportation means \cite{boccaletti2014structure, aleta2019multilayer}. Multi-layer graphs are convenient for encoding complex relationships of multiple types between data samples \cite{kivela2014multilayer}. While they can be directly tailored from a multi-relational network such as a social network data, multi-layer graphs can also be constructed from a multi-view data \cite{dong2013clustering, khasanova2016multi}, where each layer is based on one type of feature. 

In this paper, we consider data described by a multi-layer graph representation where each data sample corresponds to a vertex on the graph along with signal values acquired on each graph vertex. Each graph layer accommodates a specific type of relationship between the data samples. From a multi-view data analysis perspective, we assume that the observed signals reside on a global view, which is latent, while the information about every single view is known. Ultimately, we aim at inferring the hidden \textit{global graph} that best represents the structure of the observed signals. 

Here, the task is to employ the partial information given by the multi-layer graphs to estimate the global structure of the data.
For such a task, the connections within one layer may not have the same level of importance or multiple layers might have redundancy due to a correlation between them. Hence, it may cause information loss to consider a single layer as it is, or to merge all the layers at once \cite{magnani2013combinatorial}. In such cases, exploiting properly the information originating from each layer and combining them based on the targeted task may improve the performance of the data analysis framework.

Considering these challenges, we propose a novel technique to combine the graph layers, which has the flexibility of selecting the connections relevant to the task and dismissing the irrelevant ones from each layer. For this purpose, we employ a set of mask matrices, each corresponding to a graph layer. Through the mask combination of a priori known layers, we then learn the global structure underlying the given set of signals. The mask matrices are indicative of the contribution of each layer on the global structure. The problem of learning the unknown global graph boils down to learning the mask matrices, which is solved via an optimization problem that takes into account both the multi-layer graph representation and a signal representation model.
The signal representation model typically depends on the assumption that the signals are smooth on the unknown global graph structure.

Unlike the previous solutions learning a graph directly from a set of observations \cite{friedman2008sparse,lake2010discovering, dong2016learning, segarra2017network,pasdeloup2017characterization, egilmez2017graph}, in our study we assume that multiple graphs representing the interactions between nodes at different levels are available, and we explicitly make use of this information while learning a global graph structure.
The main benefit of the proposed method over those is that it can compensate for the often encountered case where we have a limited number of observations deviating from the assumed statistical model. Incorporation of the side information obtained from the multi-layer graph representation leads to a more reliable solution in such cases.
\begin{figure*}[h]
  \centering
  \includegraphics[width=0.8\textwidth]{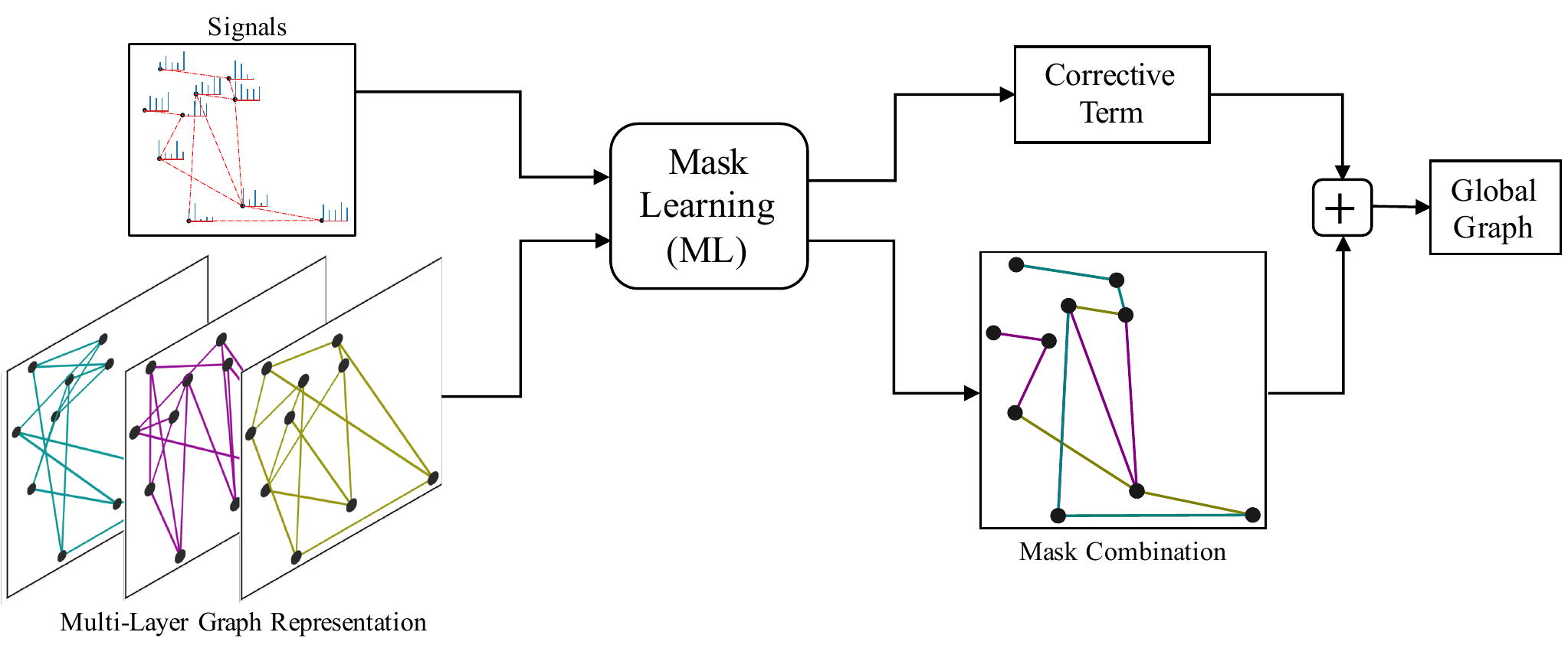}
	\caption{An illustration for the input and output of the mask learning algorithm}
  \label{fig:illust}
\end{figure*}

Fig. \ref{fig:illust} illustrates the general framework with inputs that are signals captured on a set of data samples and the multi-layer graph representation that stores the relations between those.
The set of mask matrices, which forms the mask combination of graph layers, is an output together with a corrective term bridging the gap between the multi-layer graph representation and the signal representation model. The mask combination and the corrective term are summed up to yield the global graph. 
The ultimate output is the global graph best fitting the signals.

We run experiments on a multi-relational social network dataset and a meteorological dataset where the introduced set of observations determines how to combine the multi-layer graphs into a global graph.
In the experiments on the meteorological data, for instance, we employ different types of measurements. When the type of the measurement is ``temperature", the task is to infer the global structure that well explains the temperature signals. Yet on the same set of weather stations, if we consider ``snow-fall" measurements, then the task is to infer the global structure underlying the snow-fall signals, which is found to be different from that of temperature.
The layer combination properly adapts to the target task and thus the inferred mask matrices uncover the relative importance of the layers in terms of structuring the signals of interest.
In addition, our extensive simulation results suggest that, in a structure inference problem, exploiting the additional information given by the data space through a multi-layer graph representation enhances the learning procedure by increasing its adaptability to variable input data quality.
\leavevmode\newline
\textbf{Contributions.} This paper proposes a novel structure inference framework that learns a graph structure from observations captured on a data space with partial structural information. The main contributions are summarized as follows: (i) The graph learning procedure is integrated with a multi-layer graph representation that encodes certain information offered by the data space. (ii) The task-relevant information is deduced effectively from each graph layer and combined into a global graph via a novel masking technique.
(iii) The mask matrices are optimized on the basis of the task determined by the set of observations. Hence, they indicate the relative contribution of the layers.

The rest of the paper is organized as follows. We make an overview of the related work in Section \ref{sec: Related}. In Section \ref{sec: Alg}, we present the notation used in the paper, explain the proposed algorithm and discuss it in detail. We give experimental results based on both synthetic and real-world data in Section \ref{sec: Exp}. Finally, we conclude in Section \ref{sec: Conclusion}.

\section{Related Work}
\label{sec: Related}
In this section, we present an overview of the related works i) studying the methods for the combination of multiple graphs to accomplish network analysis or semi-supervised learning tasks, ii) adopting a graph regularization framework on multi-view data for semi-supervised learning or clustering tasks, iii) constituting the state-of-the-art structure inference schemes.

In the last decade, many studies have adopted multi-layer networks to treat the data emerging in complex systems ranging from biological and technological networks to social networks, which promoted fundamental network analysis operations.
In social networks, for instance, each type of relationship between individuals may be represented by a single layer and a specific combination of the layers may reveal hidden motifs in the network. For this purpose, Magnani et al. \cite{magnani2013combinatorial} propose the concept of power-sociomatrix, which adopts all possible combinations of the layers in the analysis of a social network. Considering multiple graph representations of a data space has also gained importance in some machine learning frameworks as well. For example, Argyriou et al. \cite{argyriou2006combining} propose to adopt a convex combination of Laplacians of multiple graphs representing a data space for the semi-supervised learning task. For the same purpose, there have been also several other works studying arithmetic mean \cite{tsuda2005fast} and generalized matrix means \cite{mercado2019generalized} of multiple graphs.
From the topological perspective, such kind of combinations of multiple graphs yield the identical set of solutions of a power-sociomatrix \cite{magnani2013combinatorial} since they treat a single graph as a whole by keeping all its edges in the combination or not. Thus, they do not leave flexibility in the topology of the layer combination. In our framework, on the other hand, the masking technique has the flexibility of selecting a particular set of edges from a layer to incorporate it in the layer combination.

Moreover, many studies have employed multiple graphs in order to represent the data emerging in multi-view domains and adapted the graph regularization framework to the multi-view domain in search of a consensus of the views \cite{kumar2011co, sindhwani2005co, dong2013clustering, ioannidis2018kernel, chen2019multiview}. Since most of those studies target the semi-supervised learning or clustering tasks, a low-rank representation of the data, which is common across the views, is sufficient.
Lately, the authors in \cite{ioannidis2019recurrent} developed a Graph Neural Network scheme to conduct semi-supervised learning on data represented by multi-layer graphs, where they integrate the graph regularization approach to impose the smoothness of the label information at each graph layer.
In contrary to these methods, the proposed method specifically addresses a structure inference task which is achieved by the estimation of a graph underlying a set of observations/signals living on a multi-view/multi-layer data domain.

More recently, several graph regularization approaches have been proposed to learn a global or consensus graph from multi-view data for clustering \cite{zhan2017graph, zhou2019consensus} and semi-supervised learning \cite{li2017multi, khan2019multi}. They employ multi-view data to obtain a unified graph structure. Particularly in \cite{zhan2017graph, zhou2019consensus}, the authors propose optimization problems where single view graph representations are extracted first and then they are fused into a unified graph.
Unlike in these learning schemes, the set of observations in our settings does not belong to a specific view of the data but they are assumed to reside on an unknown global view that we aim at inferring.
In this sense, the study in \cite{khan2019multi} works in similar settings to ours. For a node classification task, it adopts a Graph Convolutional Network scheme defined on the merged graph that is obtained by adapting the method proposed in \cite{dong2013clustering}. In our case, we rather obtain the so-called global graph through a novel technique that combines the given graph layers by flexibly adapting to the structure implied by the observed signals.

The problem of learning a graph representation of the data has been addressed by various network topology inference methods. An important representative is the sparse inverse covariance estimation method via Graphical Lasso \cite{friedman2008sparse}. Later, Lake \& Tenenbaum \cite{lake2010discovering} also adopted the inverse covariance estimation approach to infer a graph Laplacian matrix. Lately, many graph learning approaches exploited the notion of smoothness \cite{dong2019learning, mateos2019connecting}. An important property of natural signals represented on graphs is the fact that they change smoothly on their graph structure. A smooth signal generative model on graphs is introduced by Dong et al. \cite{dong2016learning}, which we also adopt in our global structure inference problem in multi-layer settings.
More recently, other generative models emerged from a diffusion process are studied by \cite{pasdeloup2017characterization, segarra2017network}, where they recover a network topology from the eigenbasis of a graph shift operator such as a graph Laplacian. 
Although there are few graph learning algorithms \cite{egilmez2017graph, kalofolias2017large} allowing the incorporation of prior knowledge on the connectivity, the multi-layer domain information has not been exploited systematically in the existing structure inference approaches.
Instead, there is a line a works \cite{segarra2017joint, wang2018high, maretic2018graph, maretic2019graph} addressing the inference of multiple graphs defined on a common vertex set from a collection of observation sets, each living on one graph. Unlike those, we aim at learning a single graph, the so-called global graph, with help of a priori known multi-layer graphs that encode the additional information given by the data domain. This brings certain advantages, especially when the signal representation quality is weak due to noisy data or insufficient number of observations, where a graph learning problem is relatively ill-posed. In addition to learning the graph structure of the signals, our framework infers the contribution of different layer representations of the data to the structure of the signals.

\section{Mask Learning Algorithm}
\label{sec: Alg}
We propose a structure inference framework for a set of observations captured on a vertex space, which can be represented by multi-layer graphs.
We treat the observations captured on such a vertex space as \textit{signals} whose underlying structure is described by the hidden \textit{global graph}.
Our task is to discover the global graph by exploiting the information provided by the multi-layer graph representation and the signals.

\subsection{Multi-layer Graph Settings}
Suppose that we have $T$ graph layers, each of which stores a single type of relation between the data samples. We introduce a weighted and undirected graph to represent the relations on layer-$t$, $ \mathcal{G}_t =  (\mathcal{V}, \mathcal{E}_t, \mathbf{W}_t) $ for $t \in \{ 1,2,\cdots, T\}$, where $\mathcal{V}$ stands for the vertex set consisting of $N$ vertices shared by all the layers, and, $\mathcal{E}_t$ and $\mathbf{W}_t$ indicate the edge set and the symmetric weight matrix for layer-$t$. A graph signal $\mathbf{x} \in \mathbb{R}^N$ can be considered as a function that assigns a value to each vertex as $\mathbf{x}: \mathcal{V} \to \mathbb{R}$. We denote the set of signals defined on the vertex space $\mathcal{V}$ by a matrix $\mathbf{X} \in \mathbb{R}^{N \times K}$, which consists of $K$ signal vectors on its columns. The signals in $\mathbf{X}$ are assumed to be smooth on the unknown global graph, $ \mathcal{G} = (\mathcal{V}, \mathcal{E}, \mathbf{W}) $. The Laplacian matrix of the global graph is further given by $\mathbf{L} = \mathbf{D} - \mathbf{W}$, where $\mathbf{W}$ is the global weight matrix. $\mathbf{D}$ is the corresponding degree matrix that can be computed as
$$ \mathbf{D} = \mathrm{diag}(\mathbf{W}\mathbf{1}),$$
where $\mathbf{1}$ is the column vector of ones and $\mathrm{diag}(\cdot)$ forms a diagonal matrix from the input vector elements. $\mathbf{L}$ is a priori unknown but it belongs to the set of valid Laplacians, $\mathcal{L}$, that is composed of symmetric matrices with non-positive off-diagonal elements and zero row sum as
\begin{equation}
\resizebox{.9\hsize}{!}{$
    \mathcal{L} :=
    \left\{
      \mathbf{L} \in \mathbb{R}^{N \times N} \Bigg|
      \begin{array}{l}
        [\mathbf{L}]_{ij} = [\mathbf{L}]_{ji} \le 0,	 \forall \{(i,j): i \neq j \}\\ 
        \mathbf{L} \mathbf{1} = \mathbf{0}
      \end{array}
    \right\},
    $}
\end{equation}
where $\mathbf{0}$ is the column vector of zeros.
\subsection{Mask Combination of Layers}
Adopting the multi-layer graph and signal representation model mentioned above, we cast the problem of learning the global graph as the problem of learning the proper combination of the graph layers.
While each graph layer encodes a different type of relationship existing on the vertex space, the multiple graph layers might have some connections that are redundant or even irrelevant to the global graph structure. This requires occasional addition or removal of some edges from the layers while combining them into the global graph. For this purpose, we propose an original masking technique, which has the flexibility to properly integrate the relevant information from the layer topologies and to simultaneously adapt the global graph to the structure of the signals. We introduce the combination of layers as a masked sum of the weight matrices of the graph layers:
\begin{equation}
\label{eqn:summaskedweights}
\mathbf{W}_M = \sum_{t=1}^T \mathbf{M}_t \odot \mathbf{W}_t,
\end{equation}
where $\odot$ represents the Hadamard (element-wise) product between two matrices: the weight matrix of $\mathcal{G}_t$, which is denoted as $\mathbf{W}_t$, and the symmetric and non-negative mask matrix corresponding to layer $\mathcal{G}_t$, $\mathbf{M}_t$. The mask matrices are stacked into a variable as $\mathbf{M} = [\mathbf{M}_1 \cdots \mathbf{M}_T]$, which is eventually optimized to infer the global graph structure.
In general, the relations given in different layers may not have the same importance in the global graph. Hence, at an arbitrary edge between node-$i$ and node-$j$, the proposed algorithm learns distinct mask elements for each layer, for instance $[\mathbf{M}_t]_{ij}$ at layer $\mathcal{G}_t$ and $[\mathbf{M}_u]_{ij}$ at layer $\mathcal{G}_u$.

We finally define a function $\Lambda(\mathbf{M})$ to compute the Laplacian matrix of the mask combination given by a set of mask matrices $\mathbf{M}$ as follows:
\begin{equation}
\Lambda(\mathbf{M}) = \mathrm{diag}(\mathbf{W}_M \mathbf{1}) - \mathbf{W}_M.
\end{equation}

\subsection{Problem Formulation}
Our task now is to infer the global graph $ \mathcal{G} =  (\mathcal{V}, \mathcal{E}, \mathbf{W}) $, on which the signal set $\mathbf{X}$ has smooth variations.
Hence, in the objective function, we employ the well-known graph regularizer $\mathrm{tr}(\mathbf{X}^\intercal \mathbf{L} \mathbf{X})$, which measures the smoothness of the signal set $\mathbf{X}$ on the global graph Laplacian $\mathbf{L}$.
The optimization problem boils down to learning a set of mask matrices, $\mathbf{M}$. Within certain masking constraints, it captures the connections that are consistent with the structure of the signals from the multi-layer graph representation and yields a mask combination of the layers.
In addition, we introduce a corrective term, $\mathbf{L}_E$, which makes a transition from the mask combination obtained from the given layers to the global graph that fits the observed signals within the smooth signal representation model.
By summing it with the Laplacian of the mask combination, we express the global graph Laplacian as
$$\mathbf{L} = \Lambda(\mathbf{M}) + \mathbf{L}_E,$$
which is the ultimate output of the algorithm.
The Frobenius norm $\| \cdot \|_F$ of $\mathbf{L}_E$ permits to adjust the impact of the corrective Laplacian, $\mathbf{L}_E$, on the global graph. The overall optimization problem is finally expressed as follows:
\begin{align}
\label{eqn:objectiveE}
\begin{split}
    \operatorname*{min}_{[\mathbf{M},\mathbf{L}_E]} & \quad  \mathrm{tr}(\mathbf{X}^\intercal \Big(\Lambda(\mathbf{M})+\mathbf{L}_E\Big) \mathbf{X})  
    + \gamma \|\mathbf{L}_E \|_F^2\\
    \text{s. t.} 
    \quad & \quad [\mathbf{M}_t]_{ij} = [\mathbf{M}_t]_{ji} \ge 0, t=\{1,2,\cdots, T \}, \forall (i,j)\\
    \quad & \quad \sum_{t=1}^T [\mathbf{M}_t]_{ij} = 1,	 \forall (i,j) \\
    \quad & \quad  \Lambda(\mathbf{M}) + \mathbf{L}_E \in \mathcal{L} \\
    \quad & \quad \mathrm{tr}(\Lambda(\mathbf{M}) + \mathbf{L}_E) = \Gamma, \\
\end{split}
\end{align}
where $\gamma$ is a hyperparameter adjusting the contribution of $\mathbf{L}_E$ on $\mathbf{L}$. The last constraint on $ \mathrm{tr}(\Lambda(\mathbf{M}) + \mathbf{L}_E)$, the trace of the global graph Laplacian $\mathbf{L}$, fixes the volume of the global graph. It is set to be a non-zero value, i.e., $\Gamma > 0 $, in order to avoid the trivial solution, i.e., null global graph. It can be considered as the normalization factor fixing the sum of all the edge weights in the global graph so that the relative importance of the edges can be interpreted properly. The mask matrices are then constrained to be symmetric and non-negative, which leads to a symmetric mask combination, $\Lambda(\mathbf{M})$. The global graph Laplacian, $\mathbf{L}$, is constrained to be a valid Laplacian. Consequently, $\mathbf{L}_E$ is forced to be a symmetric matrix but it does not have to be a valid graph Laplacian matrix.
In this regard, $\mathbf{L}_E$ provides the possibility to make a subtraction from the mask combination as well as to add more weights on top of the mask combination.
We also put a constraint on the mask elements $\{[\mathbf{M}_t]_{ij}\}_{t=1}^T$, and set the search space of the mask matrices to yield a unity sum.
This establishes a dependency between the mask elements corresponding to the same edge at each layer so that the contribution of the layers at a particular connection between vertex-$i$ and vertex-$j$ is normalized. As a result of the unity sum constraint on the masks, the weight elements of the mask combination are confined into the weight range delivered by the layers as follows,
\begin{equation}
	\label{eqn:weightinterval}
    \operatorname*{min}_t [\mathbf{W}_t]_{ij} \leq [\mathbf{W}_M]_{ij} \leq \operatorname*{max}_t [\mathbf{W}_t]_{ij}.
\end{equation}
Such a restriction is actually important to keep the weight values of the global graph in a reasonable range, which is desired for the weight prediction task.
Note that dismissing an arbitrary edge $\mathcal{E}_{ij}$ from the mask combination is possible if $$\operatorname*{min}_t [\mathbf{W}_t]_{ij} = 0,$$ i.e., a connection is not defined between vertex-$i$ and vertex-$j$ in at least one of the layers.

The objective function in \eqref{eqn:objectiveE} is linear with respect to the mask matrices $\mathbf{M}$ due to the first term, and it is quadratic with respect to the corrective Laplacian $\mathbf{L}_E$ due to the second term. All the constraints are linear with respect to the optimization variables. Therefore, the problem is convex and it can be efficiently solved by quadratic programming.
\subsection{Discussion}
A theoretical analysis of the proposed problem is presented in this section, regarding the selection of the hyperparameters, the complexity and the identifiability.
\label{sec:params}

\subsubsection{Hyperparameters}
In problem \eqref{eqn:objectiveE}, we need to set two hyperparameters: $\gamma$ and $\Gamma$. First, $\gamma$ adjusts the impact of the corrective Laplacian, $\mathbf{L}_E$, on the global graph Laplacian, $\mathbf{L}$. As $\gamma$ approaches infinity, there is a full penalty on $\mathbf{L}_E$, hence the problem \eqref{eqn:objectiveE} behaves as a constrained optimization problem where $\mathbf{L}_E$ is null, i.e., $\mathbf{L}_E = \mathbf{O}$. In the other extreme case where $\gamma=0$, the global graph structure is completely defined by $\mathbf{L}_E$, which cancels out all the edges on the mask combination, $\Lambda(\mathbf{M})$, and leaves only a few edges constituting the links along which the signals are the smoothest. In this regard, $\gamma$ should be set strictly above $0$ in order to exploit the multi-layer graph representation adequately. The hyperparameter $\gamma$ is used for the purpose of interpolating the solution between the support of the multi-layer graph representation and the agreement of the signal representation. As depicted on the limit cases, the maximum exploitation of the multi-layer graph representation can be obtained when $\gamma$ approaches to infinity where the corrective term has no contribution and the global graph is directly equal to the mask combination.
Broadly speaking, $\gamma$ should be set to a high value, when the input multi-layer graph representation is more reliable than the observations. Then, smaller values should be preferred when the observations are more informative so that the mask combination is refined by the corrective term according to the agreement of the signal representation.
The factors playing a role in the quality of the input data also affect the accuracy of the proposed algorithm and they will be explained in the next part in detail.
Also, note that the value of $\gamma$ should be chosen proportionally to the squared norm of the observation matrix $\mathbf{X}$ due to the interplay between the first and the second term of the objective function in \eqref{eqn:objectiveE}. 

Second, the value of the parameter $\Gamma$ sets the volume of the global graph. Recall that the masking constraints confine the edge weights of the mask combination into the interval given by edge weights of the layers, as stated in relation \eqref{eqn:weightinterval}. Inherently, the volume of the mask combination, i.e., $\sum_{i,j}[\mathbf{W}_M]_{ij}$, is confined to the range given by the layer weight matrices. $\Gamma$ can be considered as a budget on the volume of the edges to be masked from the given graph layers together with the volume of the corrective term. Accordingly, the number of edges in the global graph is proportional to the value of $\Gamma$ as a consequence of the proposed masking approach.
For the set of solutions where $\mathbf{L}_E = \mathbf{O}$, $\Gamma$ is subject to the same feasible range for the volume of the mask combination $\mathbf{W}_M$. In that case, it has to be set as,
\begin{equation}
\label{eqn:volume}
\sum_{i,j}\operatorname*{min}_{t} [\mathbf{W}_t]_{ij} \le \Gamma \le \sum_{i,j}\operatorname*{max}_t [\mathbf{W}_t]_{ij},
\end{equation} 
so that $\mathbf{M}$ can be solved. The lower limit corresponds to the topology composed of the common edges across the layers and the upper limit corresponds to the topology given by the union of the layers. Recall that $\mathbf{L}_E$ is solved as a null matrix usually when $\gamma$ in \eqref{eqn:objectiveE} is very large, which acknowledges the full reliability on the multi-layer graphs by pushing the global graph to have the topology and the weight range provided by the layers. Decreasing the value of $\gamma$ relaxes this restriction, which enlarges the solution space for the global graph by diverting it from the mask combination solution.
To conclude, $\Gamma$ has a direct effect on the sparsity of the global graph. In practice, it can be chosen to ensure the desired sparsity level and in the feasible range of the volume of the mask combination determined by the layers as given in \eqref{eqn:volume}.

\subsubsection{Complexity Analysis}
The algorithm solves for the optimization variables consisting of the elements of the mask matrices $\{\mathbf{M}_t\}_{t=1}^T$ for $T$ layers and the elements of the corrective Laplacian matrix, $\mathbf{L}_E$. The number of optimization variables for mask elements is $O(\sum_t |\mathcal{E}_t|)$, which is the sum of the number of edges given by the layers. It can also be written as $O(ET)$, where $E$ is the average number of edges given by the layers. In the worst case, all the given layers are complete graphs where $E = \frac{N(N-1)}{2}$. However, typically, the given graph layers are sparse. If we assume that the average number of neighbors for a node in a graph layer is $k \ll N$, which makes $E = kN$, then we can say that the number of optimization variables for the mask elements grows linearly as $O(kNT)$. Second, the corrective term, $\mathbf{L}_E$, has $\frac{N(N-1)}{2}$ elements. Thus, the objective function depends on $O(N^2)$ variables quadratically and $O(kNT)$ variables linearly, which makes $O(kNT + N^2)$ in total. The number of the optimization variables has a quadratic asymptotic growth with respect to the number of nodes, $N$. It is dominated by the elements of $\mathbf{L}_E$ when $kT < N$. Moreover, due to the fact that the objective function depends quadratically on $\mathbf{L}_E$, solving for these $O(N^2)$ variables also dominates the complexity, which implies that $N$ is the factor of the complexity rather than $T$. The objective function in \eqref{eqn:objectiveE} is subject to a set of equality and inequality constraints expressed on the variables $\mathbf{M}$  and $\mathbf{L}_E$, which narrows down the solution space considerably. Ultimately, the overall complexity is determined by the quadratic programming, whose computational analysis for SDPT3 solver is given in \cite{toh1999sdpt3}.

In particular, one might desire to solve the problem in \eqref{eqn:objectiveE} in such a way that the global graph  relies entirely on the multi-layer graph representation, where the corrective term, $\mathbf{L}_E$, has no contribution.
This can be realized by choosing the hyperparameter $\gamma$ above a certain large value.
Furthermore, in certain applications, e.g., involving large networks, due to limitations on computational resources, one may also prefer to set $\mathbf{L}_E = \mathbf{O}$ and to be exempted of solving it completely.
This is possible by using a reduced version of \eqref{eqn:objectiveE} where $\mathbf{M}$ is the only optimization variable, and it is expressed by the following optimization problem:
\begin{align}
\label{eqn:ML-LP}
\begin{split}
    \operatorname*{min}_{\mathbf{M}} & \quad  \mathrm{tr}(\mathbf{X}^\intercal \Lambda(\mathbf{M}) \mathbf{X})\\
    \text{s. t.}
    \quad & \quad [\mathbf{M}_t]_{ij} = [\mathbf{M}_t]_{ji} \ge 0, t=\{1,2,\cdots, T \}, \forall (i,j)\\
    \quad & \quad \sum_{t=1}^T [\mathbf{M}_t]_{ij} = 1,	 \forall (i,j) \\
    \quad & \quad \mathrm{tr}(\Lambda(\mathbf{M})) = \Gamma, \\
\end{split}
\end{align}
which can be solved via linear programming. The objective function of this problem is equivalent to that of \eqref{eqn:objectiveE} where $\mathbf{L}_E = \mathbf{O}$. Then, the equality and inequality constraints become equivalent to those of \eqref{eqn:objectiveE} when $\mathbf{L}_E = \mathbf{O}$, noting that the first constraint in \eqref{eqn:ML-LP} already implies $\Lambda(\mathbf{M}) \in \mathcal{L}$.
Relying on these facts, we can say that uniting the solution space of \eqref{eqn:ML-LP} with $\{\mathbf{L}_E = \mathbf{O}\}$, we obtain a subset of the solution space of the problem \eqref{eqn:objectiveE}.
The reduced version requires only $O(kNT)$ optimization variables, which depends linearly on the number of nodes, $N$, hence, decreases the computational complexity considerably compared to the original problem. As a comparison, we finally note that, the unknown variable of the graph learning problems referred in Section \ref{sec: Related} is $O(N^2)$ in general.

\subsubsection{Identifiability Analysis}
The accuracy of the proposed learning scheme depends on the quality and the quantity of the input data. First of all, the factors playing a role in signal representation quality can be counted as follows:
\begin{itemize}
\item the ratio of the number of observations to the number of nodes ($K/N$). The accuracy of the statistical inference built upon the smooth signal representation model is better when there are many observations in comparison to the data dimension.
\item The signal-to-noise (SNR) of the signal set. The accuracy is better when there are clean signals that are sufficient to support the smooth signal representation model.
\item The correlation between the observations. The accuracy is better when the observations are independent and identically distributed (i.i.d.).
\end{itemize}
Note that the accuracy of the graph learning methods mentioned in Section \ref{sec: Related} are also subject to the facts above \cite{dong2019learning}.
However, theoretical guarantees of the graph Laplacian estimation methods regarding the rate of convergence and error bounds are not well explored in terms of the listed factors. Nonetheless, the graph Laplacian can be counted as a specific instance of the precision matrix of the observations \cite{dong2019learning} and there have been several works \cite{tsukuma2006improved, cai2016estimating} studying the problem of estimating normal precision matrices in more general settings.
Our algorithm estimates the graph Laplacian under quite particular priors based on a multi-layer graph structure, therefore, it is not straightforward to express a theoretical guarantee particularly fitting to our algorithm.
Yet, we argue that the benefit of the proposed learning scheme over the aforementioned graph Laplacian inference algorithms is that it does not only depend on the observations but it also profits from the information originating from the multi-layer graph representation of the data. This is advantageous especially when the signal representation quality is not fully accountable. Accordingly, the accuracy of the proposed method depends on the multi-layer graph representation quality as well. Some related parameters are:
\begin{itemize}
    \item the proportion of the global graph edges, $\mathcal{E}$, that are given by the layer edges $\mathcal{E}^L$, which can be measured by a term called \textit{coverability} introduced in \cite{magnani2013combinatorial}. Coverability is the recall of the multi-layer graph representation on the global graph, and it is calculated by $\frac{|\mathcal{E} \cap \mathcal{E}^L|}{|\mathcal{E}|}$. It measures how much the multi-layer graph representation covers the global graph and it is $1$ when the global graph is fully covered by the layers.
    \item the proportion of the common edges across the layers that are present in the global graph. The mask combination is designed to keep the intersecting edges via the mask constraints.
\end{itemize}
It is possible to relax the effect of these factors on the accuracy by choosing relatively small values of $\gamma$ and fit the global graph more with respect to the information emerging from the observations.

\section{Experiments}
\label{sec: Exp}
We compare the global graph recovery performance of our method (\textbf{ML}) against some state-of-the-art graph learning algorithms. First, we compare the graph learning algorithm that we consider as baseline \cite{dong2016learning}, which is referred to as \textbf{GL-SigRep}. To make a fair assessment, we compare our method to another version of \textbf{GL-SigRep}, where the graph learning algorithm is informed of the input layers by restricting its solution space to the set of edges given by the layers as below:
\begin{align}
\label{eqn:GLinformed}
\begin{split}
    \operatorname*{min}_{\mathbf{L}} & \quad  \mathrm{tr}(\mathbf{X}^\intercal \mathbf{L} \mathbf{X}) + \gamma\|\mathbf{L}\|_F\\
    \text{s. t.} 
    \quad & \quad  \mathbf{L} \in \mathcal{L} \\
    \quad & \quad \mathrm{tr}(\mathbf{L}) = N \\
    \quad & \quad [\mathbf{L}]_{ij} = 0, \text{ for } \{(i,j): [\mathbf{W}_t]_{ij} = 0, \forall t\}.
\end{split}
\end{align}
We refer to this method as \textbf{GL-informed}.

We also compare against the optimal convex combination of the layers. For that purpose, we adapt the method for learning the convex combination of multiple graph Laplacians introduced in \cite{argyriou2006combining} for our settings as in the following optimization problem:
\begin{align}
\label{eqn:GLconv}
\begin{split}
    \operatorname*{min}_{\boldsymbol\alpha} & \quad   \mathrm{tr}(\mathbf{X}^\intercal \mathbf{L} \mathbf{X})  +  \beta \| \boldsymbol\alpha \|_2^2
     \\
    \text{s. t.} 
    \quad & \quad \mathbf{L} = \sum_{t=1}^T \alpha_t \mathbf{L}_t \\
    \quad & \quad \alpha_t \ge 0, \forall t \\
    \quad & \quad \sum_{t=1}^T \alpha_t = 1,
\end{split}
\end{align}
where the coefficients $\boldsymbol\alpha = [\alpha_1 \cdots \alpha_T]$ are learned for the convex combination of the layer Laplacians, $\{\mathbf{L}_t\}_{t=1}^T$, to reach the global graph Laplacian $\mathbf{L}$. Throughout this section, the algorithm solving the problem \eqref{eqn:GLconv} is referred to as \textbf{GL-conv}.

For the quantitative assessment of link prediction performance, we employ the following evaluation metrics: \textit{Precision, Recall} and \textit{F-score} \cite{manning2010introduction}. We also compute the \textit{mean squared error (MSE)} of the inferred weight matrix for the assessment of weight prediction performance. We solve the problems \textbf{ML} \eqref{eqn:objectiveE}, \textbf{GL-informed} \eqref{eqn:GLinformed}, \textbf{GL-SigRep} \cite{dong2016learning} and \textbf{GL-conv} \eqref{eqn:GLconv} via quadratic programming for which we utilize the CVX toolbox \cite{grant2014cvx} with SDPT3 and MOSEK \cite{mosek} solver and the code is available online\footnote{https://github.com/bayrameda/MaskLearning}.
\begin{figure*}[!h]
  \captionsetup{justification=centering}
  \hspace*{0.5cm}
  \includegraphics[width=0.95\textwidth]{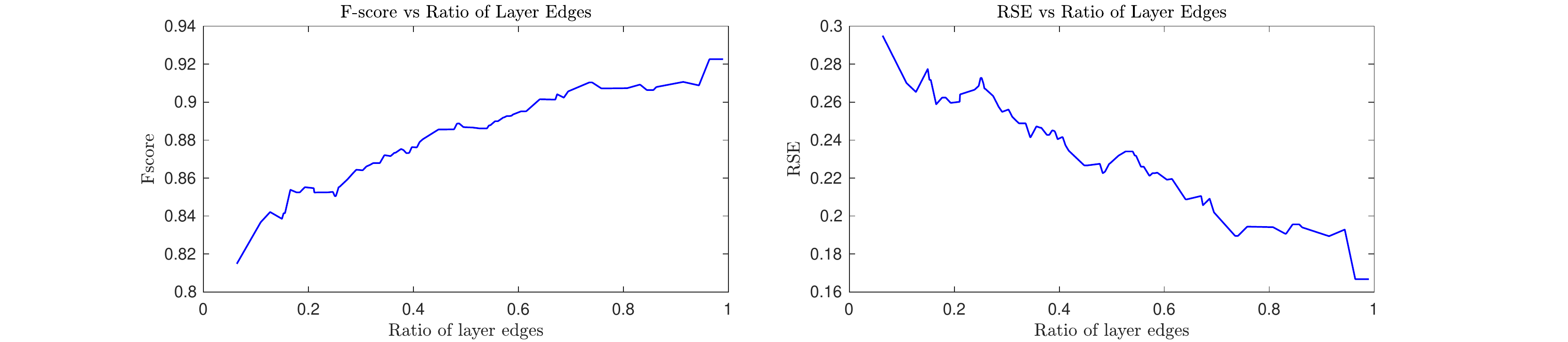}
	\caption{Performance with respect to the ratio of layer edges}
  \label{fig:ratio}
\end{figure*}
\subsection{Experiments on Synthetic Data}
In this section, we run experiments on two different scenarios. First, we generate the global graph in a fully complementary scenario where the mask combination of the layers is directly equal to the global graph. Second, we test the algorithms on a non-fully complementary scenario where the global graph is created from a perturbation on the topology of the mask combination. For both cases, we generate the mask combination and the signal set as follows:
\leavevmode\newline
\textbf{Generation of layers and the mask combination.} First, the vertex set 
$\mathcal{V}$ is established with $|\mathcal{V}| = N$ vertices whose coordinates are generated randomly on the 2D unit square with a uniform distribution. Next, an edge set $\mathcal{E}^L$ is constructed for the layers by putting edges between all pairs of vertices in $\mathcal{V}$ whose Euclidean distance is under a certain threshold. The edge weights are computed by applying a Gaussian kernel, i.e., $\mathrm{exp}(-d(i,j)^2/ 2\sigma^2)$, where $d(i,j)$ is the distance between vertex-$i$ and vertex-$j$ and $\sigma = 0.45$.
To generate two graph layers, $\mathcal{V}$ is randomly separated into two neighborhood groups: $\mathcal{V}_1$ and $\mathcal{V}_2$.
Let us denote the set of edges connecting the vertices in one group $\mathcal{V}_t$, to all vertices in $\mathcal{V}$ as $\mathcal{E}^L_{\mathcal{V}_t, \mathcal{V}}$. The graph layer $\mathcal{G}_t$ is built on the edge set $\mathcal{E}_t = \mathcal{E}^L_{\mathcal{V}_t, \mathcal{V}}$, and the corresponding edge weights are used to construct its weight matrix $\mathbf{W}_t$.
For the generation of the masks, another set of edges $\mathcal{E}^M$, a subset of $\mathcal{E}^L$ whose edge weights are above $\tau = 0.8$, are reserved.
Let us denote the set of edges in $\mathcal{E}^M$ that are between a pair of vertices in $\mathcal{V}_t$ as $\mathcal{E}^M_{\mathcal{V}_t, \mathcal{V}_t}$.
The mask matrix $\mathbf{M}_t$ is constructed by setting its entries corresponding to the edges in $\mathcal{E}^M_{\mathcal{V}_t, \mathcal{V}_t}$ as $1$.
Also, all the entries corresponding to the common edges between the layers, $\mathcal{E}_1 \cap \mathcal{E}_2$, are set as $0.5$ in the mask matrices in order to keep the intersection of the layers in the mask combination.
Lastly, the weight matrix of the mask combination is computed via the formulation given in \eqref{eqn:summaskedweights}. As the next step, the global graph is produced according to one of the experimental scenarios that will be explained in the following sections.
\leavevmode\newline
\textbf{Signal Generation.} Following the generation of the mask combination and the global graph, the global graph Laplacian matrix, $\mathbf{L}$, is computed. Using that, a number of smooth signals are generated according to the generative model introduced in \cite{dong2016learning}. Basically, the graph Fourier coefficients $\mathbf{h}$ of a sample signal can be drawn from the following distribution;
\begin{equation}
    \mathbf{h} \sim \mathcal{N}(0,\mathbf{\Sigma})
\end{equation}
where $\mathbf{\Sigma}$ is the Moore-Penrose pseudo-inverse of $\mathbf{\Sigma}^\dagger$, which is set as the diagonal eigenvalue matrix of $\mathbf{L}$.
The eigenvalues, which are associated with the main frequencies of the graph, are sorted in the main diagonal of $\mathbf{\Sigma}^\dagger$ in ascending order. Thus, the signal Fourier coefficients corresponding to the low-frequency components are selected from a normal distribution with a large variance while the variance of the coefficients decreases towards the high-frequency components. In other words, the signal is produced to have most of its energy in the low frequencies, which enforces smooth variations in the expected signal over the graph structure. A signal vector is then calculated from $\mathbf{h}$ through the inverse graph Fourier transform \cite{ortega2018graph}.
\subsubsection{Fully Complementary Scenario}
We first conduct experiments where the global graph is directly equal to the mask combination. We refer to this data generation setting as the fully-complementary scenario since the edge set of the global graph is fully covered by the union of the layers, thus, the coverability is fixed to 1. We generate 50 smooth signals on the global graph. Its volume is normalized by the number of vertices, $N = 20$. \textbf{GL-informed} \eqref{eqn:GLinformed} already learns a graph with a volume of $N$, therefore, we set the parameter $\Gamma = N$ in \textbf{ML} as well. The volume of the graph learned by \textbf{GL-conv} \eqref{eqn:GLconv} is also normalized to $N$ for a fair comparison of the MSE score. This experimental scenario (generation of fully complementary layers, global graph and signal set) is repeated 20 times and the performance metrics are averaged on these 20 instances. The findings are summarized in Table \ref{tbl: full}. Following the discussion in Section \ref{sec:params}, we employ the reduced version of \textbf{ML} in \eqref{eqn:ML-LP}, since the corrective term $\mathbf{L}_E$ is not required in the fully-complementary settings. Consequently, the global graph is inferred to be directly equal to the mask combination.
In Table \ref{tbl: full}, \textbf{GL-conv} yields a high difference between the recall and the precision rate since it either picks the edge set of a layer as a whole or not. Therefore, it is not able to realize an edge-specific selection, which leads to poor F-score compared to other methods. The global graph recovery performance of \textbf{GL-informed} is presented as a surrogate of \textbf{GL-SigRep}, since the solution for the global graph already lies in the edge set given by the layers in fully-complementary settings. The MSE score of \textbf{ML} and \textbf{GL-conv} is better than the one of \textbf{GL-informed}. This is due to the fact that \textbf{ML} and \textbf{GL-conv} have better guidance on the weight prediction task by confining the interval of weight values of the global graph to the interval introduced by the layers, which is expressed in \eqref{eqn:weightinterval} for \textbf{ML}.
\begin{table}[h]
\caption{Global Graph Recovery and Mask Recovery Performances}
\label{tbl: full}
\resizebox{\columnwidth}{!}{%
\begin{tabular}{cc||cccc}
                                                &             & \textbf{precision} & \textbf{recall} & \textbf{F-score} & \textbf{MSE} \\
                                                \hline
\multicolumn{1}{p{1.1cm}|} {\multirow{2}{*}{\parbox{1.2cm}{\textbf{Global Graph Recovery}}}} & {\textbf{ML}} & 86.98\% & 90.79\% & \textbf{88.84}\% & \textbf{1.6E-03}   \\
\multicolumn{1}{p{1.1cm}|}{}             & \textbf{GL-informed} & 81.26\% & 88.91\% & 84.48\% & 2.6E-03\\  \multicolumn{1}{p{1.1cm}|}{} &\textbf{GL-conv} & 63.82\% & 100\% & 77.41\% & 2.1E-03\\
                                                \hline
\multicolumn{1}{p{1.1cm}|}{\textbf{Mask \newline Recovery}}  & \textbf{ML} & 92.57\% & 94.88\% & 93.68\% &  - 
\end{tabular}
}
\end{table}
Finally, \textbf{ML} achieves good rates on the mask recovery performance, which measures how correctly the algorithm selects the edges from each layer to form the mask combination.

In this setting, in each repetition of the experiment, the number of edges given by each layer is also recorded to see the effect of the ratio of the layer edges e.g., $|\mathcal{E}_1|/|\mathcal{E}_2|$, on the performance of \textbf{ML}. Note that, $|\mathcal{E}_1|/|\mathcal{E}_2| = 1$ means that the layers are completely balanced and $|\mathcal{E}_1|/|\mathcal{E}_2| = 0$ means that one layer is completely deficient in terms of the number of edges. Here, we employ the \textit{relative squared error} (RSE) as a metric to assess the change in the accuracy of weight matrix estimation, which is the normalized form of the squared error by the squared norm of the ground truth weight matrix.
In Fig. \ref{fig:ratio}, we see that the performance is enhanced approximately by $12\%$ when the layers are balanced compared to the deficient layer case. 
We argue that the reason for such an enhancement is the improvement in the alignment between the layers, considering the fact that the masking coefficients are constrained in a way to keep the intersecting edges between the layers in the mask combination. In other words, when the layers are more balanced, there is a higher chance of a larger intersection. Hence, we speculate that balanced layers may lead to better performance as long as the alignment between the layers is important for the structure of the observations, as in the case of the synthetic data generated in fully complementary settings.
In the extreme case where the intersection is empty when one layer is completely deficient, the performance obviously loses the gain that could be obtained from the overlap between the layers.
\begin{figure}[!h]
  \centering
  \includegraphics[width=0.5\textwidth]{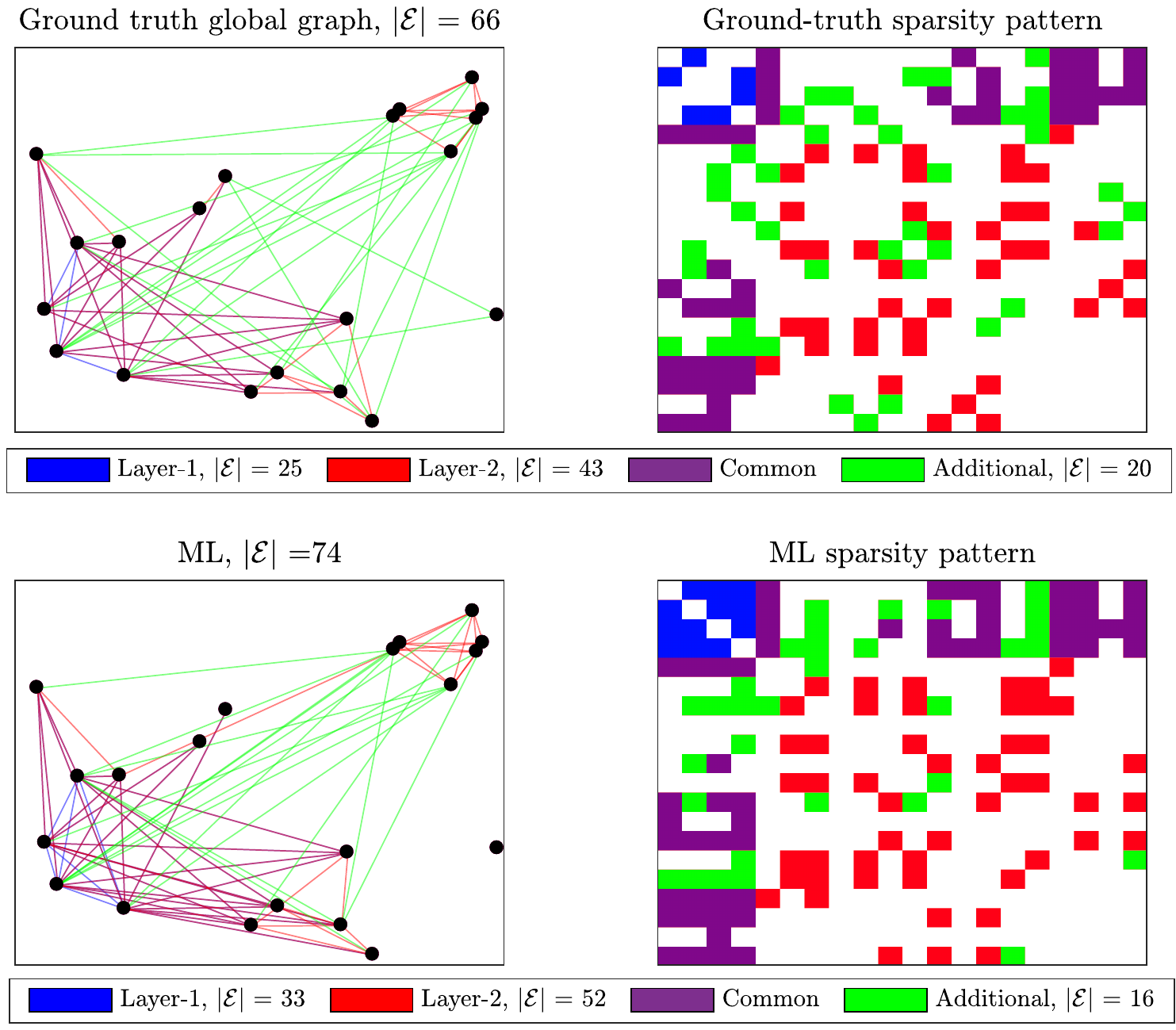}
	\caption{Ground truth global graph and the solution given by \textbf{ML}}
  \label{fig:toy}
\end{figure}
\subsubsection{Non-fully complementary scenario}
In this section, we test the algorithms in experiments where the data is generated with different levels of multi-layer and signal representation quality so that we analyze their effects on the global graph recovery performance.
First, to create the global graph, we deviate from the exact mask combination by perturbing its topology to some degree. Basically, we randomly replace a set of edges existing on the mask combination outside the union of the graph layers. The degree of such a perturbation on the mask combination can be measured by the coverability. The larger the number of edges perturbed on the topology of the mask combination, the more the global graph diverts from the multi-layer graph representation, which decreases the coverability. Consequently, the multi-layer representation quality drops. A demonstration is provided in Fig. \ref{fig:toy} top row where the global graph is generated with coverability 0.7. Here, the set of edges outside the mask combination is shown in green. As seen in Fig. \ref{fig:toy} bottom row, \textbf{ML} manages to predict some edges that are not given by the multi-layer graph representation, thanks to the corrective term in \eqref{eqn:objectiveE}.
\begin{figure*}[h]
  \captionsetup{justification=centering}
  \hspace*{2cm}
  \includegraphics[width=0.8\textwidth]{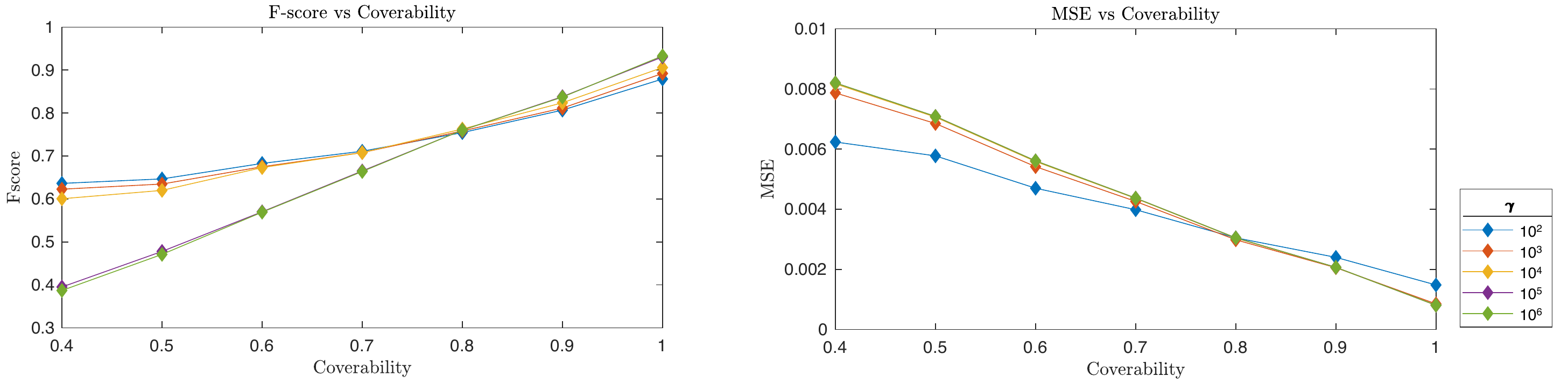}
	\caption{Performance of \textbf{ML} with different $\gamma$ values vs coverability}
  \label{fig:ml_gamma}
\end{figure*}
\begin{figure*}[!h]
  \captionsetup{justification=centering}
  \hspace*{1.5cm}
  \includegraphics[width=0.9\textwidth]{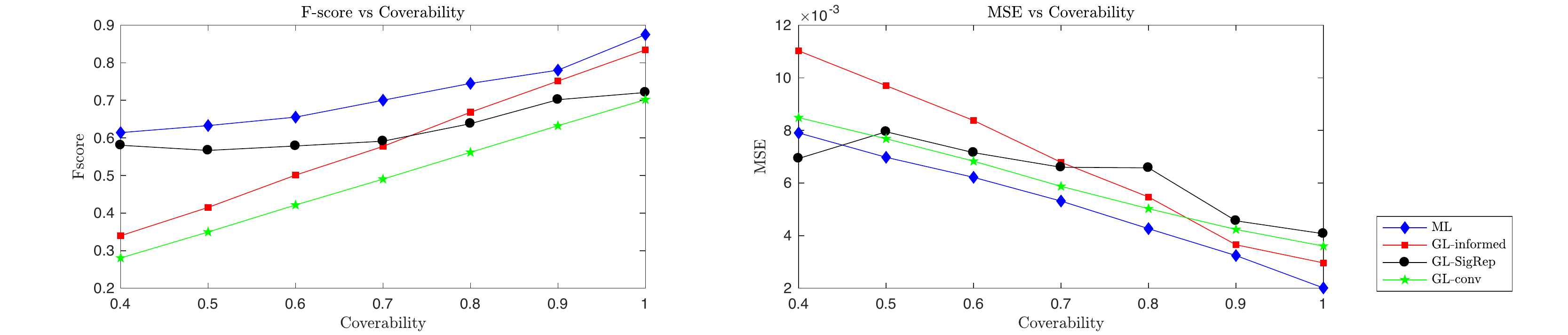}
	\caption{Performance of the algorithms vs coverability}
  \label{fig:coverability}
\end{figure*}
\leavevmode\newline \textbf{Effect of multi-layer representation quality.} Here, we test the performance of \textbf{ML} in non-complementary settings with different coverability and different values of $\gamma$.
We conduct each experiment with signal sets composed of 50 signals that are generated on the global graph as explained before.
We average the performance metrics on 20 experiments in Fig. \ref{fig:ml_gamma}.
The following observations can be made: (i) When coverability has the lowest value (0.4), \textbf{ML} with $\gamma = 100$ has the best performance. (ii) When it has the highest value (1), which corresponds to fully complementary settings, \textbf{ML} with $\gamma = 10^6$ has the best performance. (iii) Whatever value is chosen for the parameter $\gamma$, the performance of \textbf{ML} gets better with increasing coverability.  Considering these facts, choosing a smaller value for the parameter $\gamma$ seems to be a good remedy for lower coverability settings. Yet, this degrades the performance slightly in the high coverability settings, which confirms the theoretical analysis given in \ref{sec:params}. Hence, if there is no prior knowledge on the reliability of the multi-layer graph representation or the signal representation, one may prefer to use small values for $\gamma$ by compromising a small decay in the performance in the case of highly reliable multi-layer graph representation.
Moreover, the performance of \textbf{ML} improves as the global graph approaches the mask combination of the layers. This is simply because the algorithm bases the global graph on top of the mask combination, and any modification made on it by the corrective term is subject to an extra cost and thus limited.
Therefore, \textbf{ML} with any $\gamma$ value performs best when the mask combination is directly equal to the global graph, which is possible only in the fully complementary settings. Still, the corrective term improves the performance in the non-fully complementary settings. 
Given the plots in Fig. \ref{fig:ml_gamma}, an appropriate $\gamma$ value for each coverability interval can further be found. For example, it can be chosen as $\gamma = 100$ for coverability $\le 0.75$, then $\gamma = 10^4$ until coverability $= 0.8$, $\gamma = 10^5$ later until coverability $= 0.9$ and $\gamma = 10^6$ for coverability $> 0.9$.
We now adopt these values to present the performance of \textbf{ML} against the competitor algorithms by averaging the performance metrics on 20 experiments in each coverability setting, given in Fig. \ref{fig:coverability}. Beginning with the performance of \textbf{GL-informed}, we see that its performance improves regularly with the raising coverability, and it outperforms \textbf{GL-SigRep} for coverability $\ge 0.73$. The coverability is irrelevant for the performance of \textbf{GL-SigRep} since it receives no multi-layer guidance, hence the fluctuations can be disregarded as the coverability changes. Nonetheless, its performance slightly drops in low coverability settings. This is because the edges of the global graph are rewired randomly outside the union of the layers , which renders the graph towards a random network. It is acknowledged in \cite{dong2016learning} that graph learning from smooth signals in random network structures has slightly lower performance than learning on regular networks. Still, in Fig. \ref{fig:coverability}, the performance of \textbf{GL-SigRep} in black line should be considered as a reference since it is the least affected by the coverability. Furthermore, the trend of \textbf{ML} in blue line seems to be more resistant than \textbf{GL-informed} in low coverability settings, thanks to the corrective term. The performance of  \textbf{ML} approaches \textbf{GL-SigRep} as coverability decreases since the multi-layer guidance diminishes. Yet, it manages to keep its F-score above \textbf{GL-SigRep} even where the coverability is low. The MSE of \textbf{GL-conv} follows a similar path with \textbf{ML}. Yet, \textbf{ML} achieves a lower MSE due to the flexibility in the edge selection process and the corrective term. The F-score of \textbf{GL-conv}, on the other hand, is inferior compared to the other methods since it simply merges the topology of the layers without an edge selection process.
\begin{figure*}[!h]
  \captionsetup{justification=centering}
  \hspace*{1.5cm}
  \includegraphics[width=0.9\textwidth]{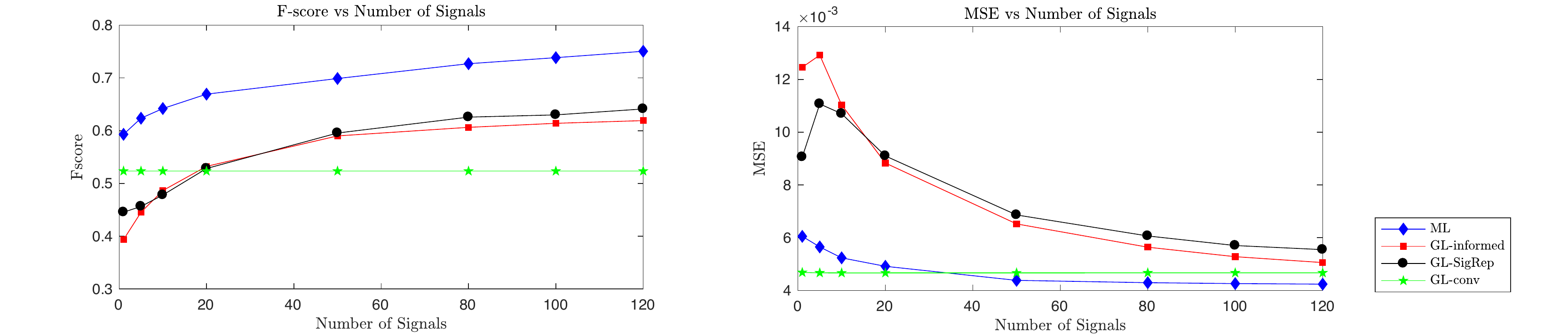}
	\caption{Performance of the algorithms vs number of signals}
  \label{fig:numsig}
\end{figure*}
\begin{figure*}[!h]
  \captionsetup{justification=centering}
  \hspace*{1.5cm}
  \includegraphics[width=0.9\textwidth]{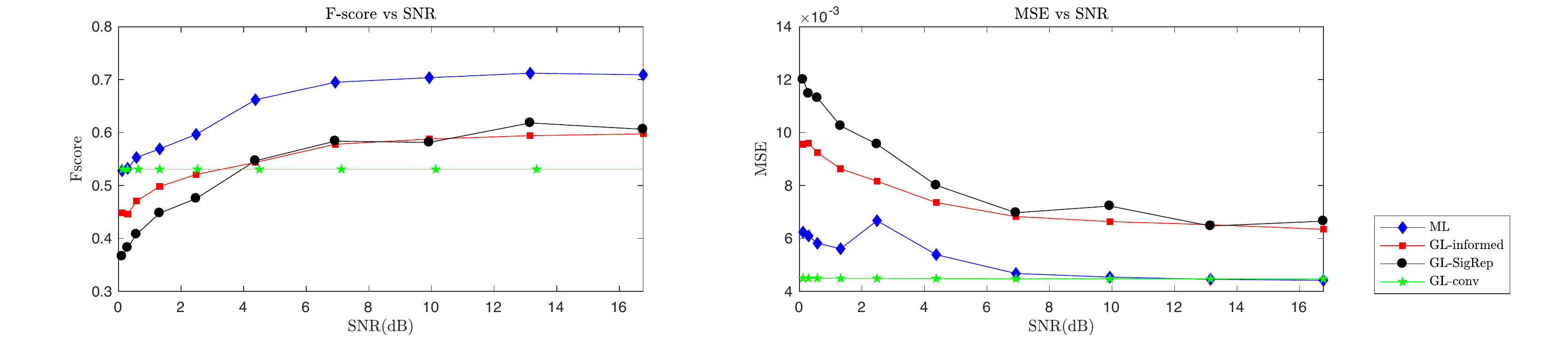}
	\caption{Performance of the algorithms vs signal quality}
  \label{fig:noise}
\end{figure*}
\leavevmode\newline \textbf{Effect of signal representation quality.}
Here, we use a fixed coverability of 0.7 to generate the global graph and the parameter $\gamma$ for \textbf{ML} is set to $100$. We first evaluate the global graph recovery of the algorithms by generating different numbers of signals on the global graph. The findings are averaged on 20 different instances of this scenario and plotted in Fig. \ref{fig:numsig}.
Then, we measure the performance of the algorithms on signal sets with different SNR values, which is given in Fig. \ref{fig:noise}. To do that, we generate a noise content from normal distribution at random with different variance values and add it to the signal set. As expected, all the methods but \textbf{GL-conv} achieve better performance as the number of signals increases, or, as the noise power drops.
\textbf{GL-conv}, on the other hand, is the least affected by the changes in the number of signals. The strictness of the convex combination constraint permits to obtain a similar combination even when there are few signals or noisy signals. Yet, this further prevents enhancing its performance in the high signal representation quality conditions. For instance in Fig. \ref{fig:numsig}, \textbf{ML} achieves a lower MSE than \textbf{GL-conv} when there is a high number of signals.
Based on the plots in Fig. \ref{fig:coverability}, it is already known that around $70\%$ coverability, \textbf{ML} achieves a good performance that is followed by \textbf{GL-SigRep} and \textbf{GL-informed}. This is also confirmed by the plots in Fig. \ref{fig:numsig} and \ref{fig:noise}. \textbf{GL-SigRep} is the method that is the most affected by the signal quality since it is not able to compensate for the lack of knowledge in the signal set. On the other hand, \textbf{ML} is resistant to the change in the signal quality, since it exploits the multi-layer guidance. In addition, \textbf{ML} permits flexibility in the learning scheme by adjusting the $\gamma$ parameter according to the signal quality. For example, in Fig. \ref{fig:noise}, under $2$dB SNR, we use $\gamma = 10^7$, so that the learning process relies more on the multi-layer graph representation. Therefore, \textbf{ML} manages to perform better than the competitor algorithms in low SNR conditions.
\subsection{Learning from Meteorological Data}
We now present experiments on real datasets and focus first on the meteorological data provided by Swiss Federal Office of Meteorology and Climatology (MeteoSwiss)\footnote{https://www.meteoswiss.admin.ch/home/climate/swiss-climate-in-detail/climate-normals/normal-values-per-measured-parameter.html}. The dataset is a compilation of 17 types of measurements including temperature, snowfall, precipitation, humidity, sunshine duration, recorded in weather stations distributed over Switzerland. Monthly normals and yearly averages of the measurements calculated based on the time period 1981-2010 are available at 91 stations.
For the stations, we are also provided geographical locations in GPS format and altitude values, i.e., meters above sea level. We use each type of measurement as a different set of observations to feed the graph learning framework. Our goal is to explain the similarity pattern for each type of measurement with the help of geographical location and altitude of the stations. 
\leavevmode\newline \textbf{Multi-Layer Graph Representation.}
We construct a 2-layer graph representation where vertices are the stations, which are connected based on GPS proximity in one layer and based on altitude proximity in the other one. We construct the layers as unweighted graphs by inserting an edge between two stations that have Euclidean distance below a threshold, which is set to an edge sparsity level of $10 \%$. Consequently, each graph layer has approximately the same number of edges so that the edge selection process during mask learning is not biased by any layer. We normalize the adjacency matrices of the layers to fix the volume of the graph layers to the number of vertices, $N$, which is also used as the value of the parameter $\Gamma$ in \textbf{ML}.
\subsubsection{Learning Masks from Different Set of Measurements}
\begin{figure*}[!h]
\begin{minipage}[b]{0.48\linewidth}
  \centerline{\includegraphics[width=1\linewidth]{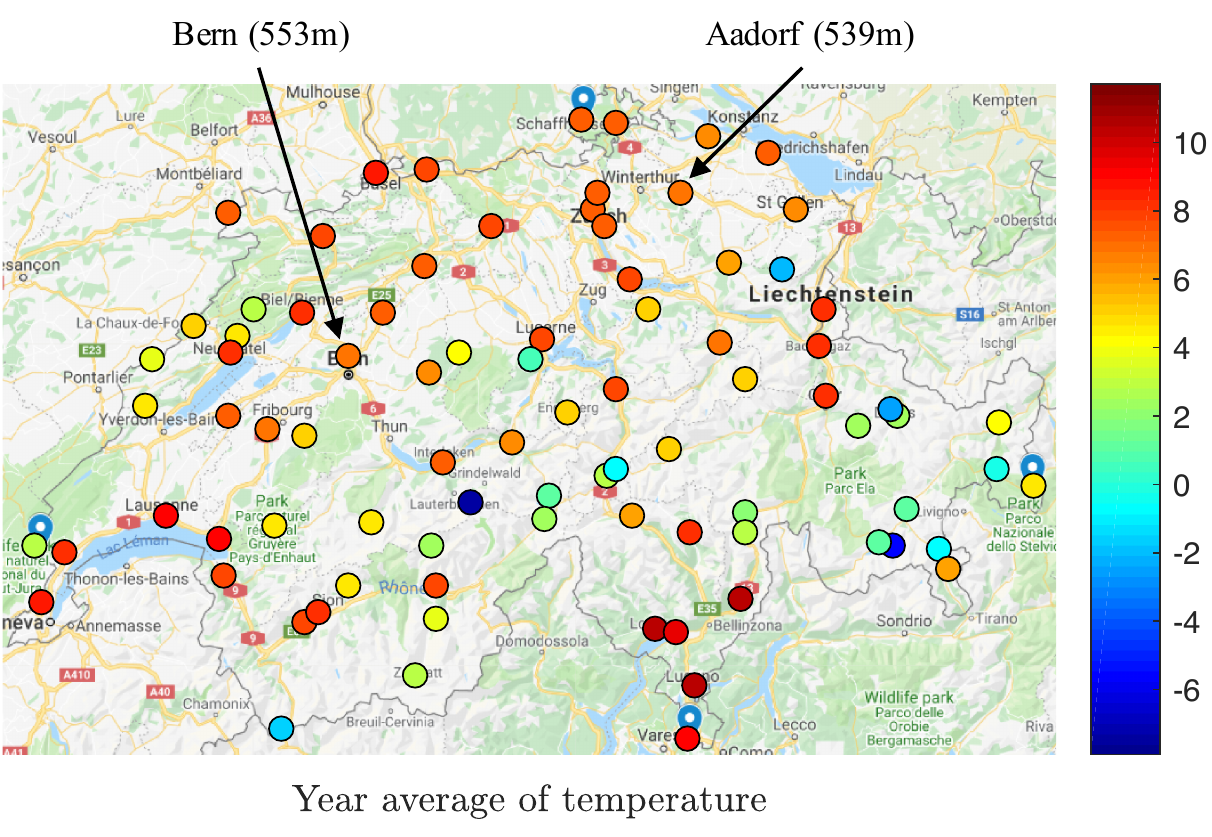}}
\end{minipage}
\begin{minipage}[b]{0.48\linewidth}
  \centerline{\includegraphics[width=1\linewidth]{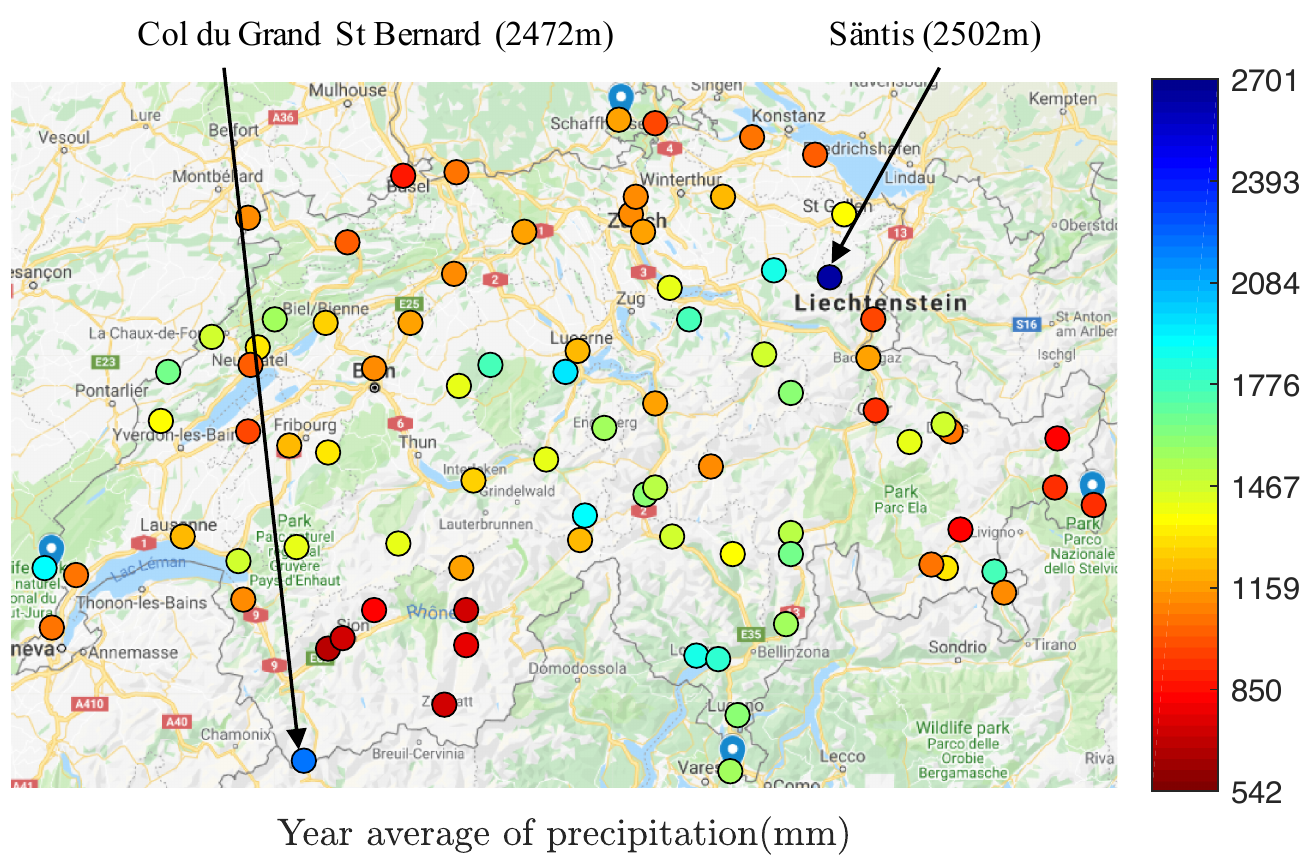}}
\end{minipage}
\caption{Year average of temperature and precipitation}
\label{fig:Map}
\end{figure*}
We test the mask learning algorithm on different types of observations separately. We use the monthly normal of the measurements as the signal set, which makes the number of signals $K =12$. Here, the yearly averages are not used for graph learning, instead, they will be used for a visual assessment of the learned graph.
We assume that the similarity between the measurement patterns of two stations must be explained either by geographical proximity or elevation similarity. Due to this, we employ \textbf{ML} in the reduced version \eqref{eqn:ML-LP} to learn a global graph structure with the fully complementary assumption.
It is possible to interpret the significance of the geographical location proximity and the altitude proximity in the formation of each type of observation by examining the mask matrices inferred by \textbf{ML}.
\begin{table}[h]
\centering
\caption{Contribution of layers on the structure of different measurements}
\label{tbl: GpsAlt}
\begin{tabular}{l|cc}
Measurement        & GPS  & Altitude \\
\hline
Temperature        & 36\% & 64\%     \\
Snowfall (cm)      & 37\% & 63\%     \\
Humidity           & 51\% & 49\%     \\
Precipitation (mm) & 52\% & 48\%     \\
Cloudy days        & 65\% & 35\%     \\
Sunshine (h)       & 54\% & 46\%    
\end{tabular}
\end{table}

In Table \ref{tbl: GpsAlt}, the percentage of the connections that \textbf{ML} draws from the GPS and the altitude layer is given for different types of measurements that are used as signals. To begin with temperature, its structure seems to be highly coherent with the altitude similarity considering the percentage contribution of each layer.
We further check the yearly temperature averages, which is shown in Fig. \ref{fig:Map}. According to that, Bern and Aadorf are the stations providing the most similar average. Indeed, an edge is inferred between them on the global structure of the temperature measurements, and it is extracted from the altitude layer where the two stations are connected within 14m elevation distance.
The correlation between temperature measurements and altitude is also noted by the authors in \cite{dong2016learning}. Similar to temperature, snowfall is also anticipated to be highly correlated with the altitude of the stations. This is also what is derived by \textbf{ML} which draws more connections from the altitude layer than the GPS layer as given in Table \ref{tbl: GpsAlt}. The `cloudy days' measurement, however, is found to be highly coherent with the GPS proximity by drawing $65\%$ of its connections from the GPS layer. Next, humidity, precipitation and sunshine are evenly correlated with both of the GPS and altitude layers, according to Table \ref{tbl: GpsAlt}.
Given the yearly average of precipitation shown in Fig. \ref{fig:Map}, Geneva and Nyon have the closest records. As seen, they are also pretty close on the map and thus their connection on the global graph of precipitation is drawn from the GPS layer. In addition, Fey and Sion are the stations providing the lowest records on average, and their connection is also drawn from the GPS layer. On the other hand, Col du Grand-Saint-Bernard and S\"{a}ntis display the highest records, and they are connected in the altitude layer with 30m elevation distance between them.
\begin{figure}[!h]
  \centering
  \captionsetup{justification=centering}
  \includegraphics[width=0.5\textwidth]{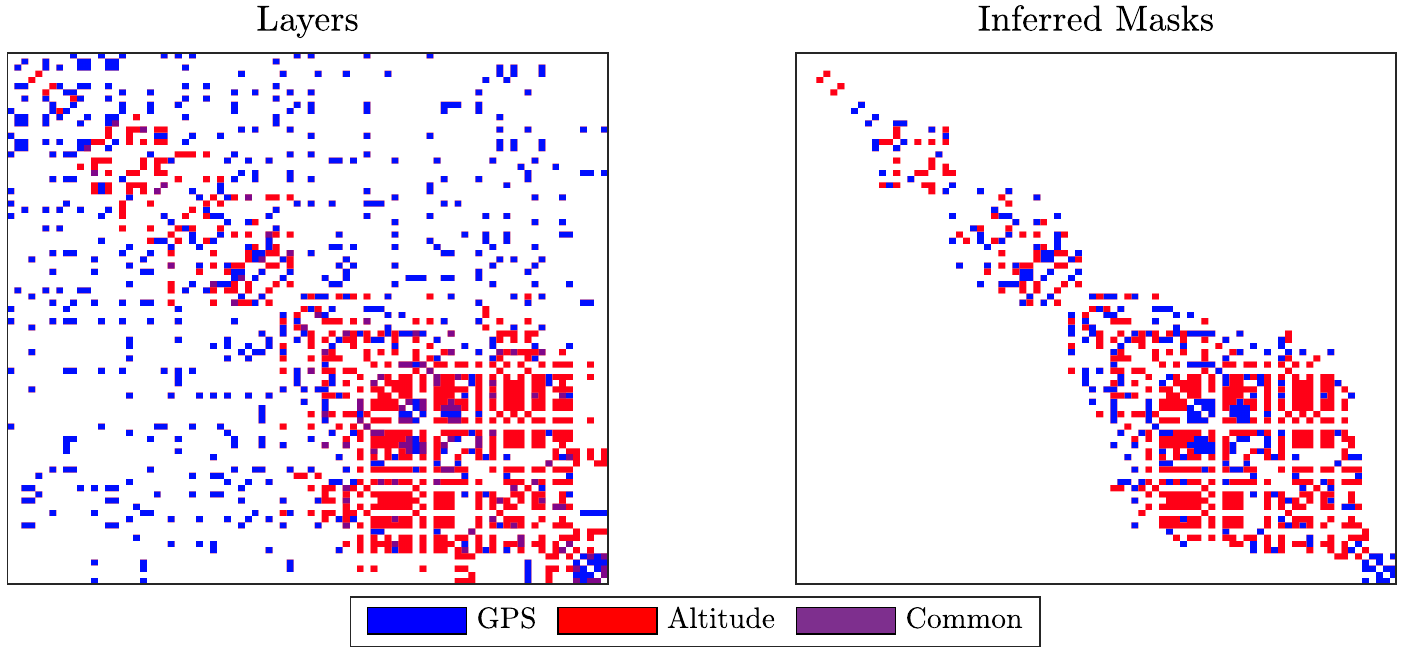}
	\caption{Sparsity pattern of the layers and the masks with respect to year average of temperature}
  \label{fig:temp_sparse}
\end{figure}

Furthermore, in Fig. \ref{fig:temp_sparse}, we visualize the layer adjacency matrices and the inferred mask matrices by sorting the vertices (stations) with respect to their yearly average temperature measurements. Recall from Table \ref{tbl: GpsAlt} that the altitude layer is found to be dominant for explaining similarities in temperature. This is also evident by the connectivity pattern of the layers, which is shown on the left of Fig. \ref{fig:temp_sparse}. The GPS layer connectivity is distributed broadly whereas the altitude layer connections are gathered around the main diagonal, which contains the edges between the vertices that are similar in yearly average.
On the right of Fig. \ref{fig:temp_sparse}, we see that inferred mask matrices for both of the layers are organized along the diagonal. This indicates that the algorithm manages to dismiss the connections that are irrelevant to the similarity pattern of temperature, especially on the GPS layer.
\subsubsection{Signal Inpainting on the Global Graph}
We now prepare a signal inpainting experiment to point out the benefits of learning a proper global graph representation.
We consider the monthly normals of the temperature measurements as the signal set. The vertex set is composed of $86$ stations that are providing temperature measurements, i.e., $N=86$. Then, a graph structure is inferred from those observations using \textbf{GL-SigRep}. In addition, by taking the multi-layer graph representation into account, a global graph structure is inferred using \textbf{GL-informed}, \textbf{GL-conv} and \textbf{ML}. During the graph learning process, we train the algorithms by the measurements on $11$ months and then try to infer the measurements of the remaining month via inpainting. In the inpainting task, we remove the values of the signal on half of the vertices selected randomly. Our aim is to recover the signal values on the whole vertex space by leveraging the known signal values and the learned graph. We solve the following graph signal inpainting problem \cite{perraudin2017stationary}:
\begin{equation}
\label{eqn:inpainting}
\operatorname*{min}_{\mathbf{x}} \quad   \|\mathbf{S}\mathbf{x} - \mathbf{y}\|_2^2   + \gamma (\mathbf{x}^\intercal \mathbf{L} \mathbf{x}),
\end{equation}
which has a closed form solution as:
\begin{equation}
\label{eqn:inpainting_closed}
\mathbf{x} = (\mathbf{S}^\intercal \mathbf{S} +  \gamma \mathbf{L})^{-1}\mathbf{S}^\intercal \mathbf{y},
\end{equation}
where $\mathbf{y} \in \mathbb{R}^l$ is the vector containing the known signal values by the algorithms, and $\mathbf{x} \in \mathbb{R}^N$ is the vector that contains the recovered signal values on all the vertices. $\mathbf{S} \in \mathbb{R}^{l \times N}$ is a mapping matrix reducing $\mathbf{x}$ to a vector whose entries correspond to the vertex set with the known signal values. Therefore, $\mathbf{S}^\intercal \mathbf{S} $ is a diagonal matrix whose non-zero entries correspond to this vertex set.

We repeat the graph learning and inpainting sequence on $12$ instances where the number of signals used in the graph learning part is $K=11$ and the inpainting is conducted on the values of a different month at each time. We calculate the MSE between the original signal vector and the recovered signal vector. In addition, we compute the \textit{mean absolute percentage error (MAPE)}, which measures the relative absolute error with respect to the original signal magnitudes. We average the performance metrics over $12$ instances for each algorithm used in the graph learning part, which is given in Table \ref{tbl: inpaint}.
\begin{table}[h]
\centering
\caption{Signal inpainting performance of the algorithms}
\label{tbl: inpaint}
\begin{tabular}{c|cc|}
\cline{2-3}
            & \textbf{MSE}   & \textbf{MAPE}   \\
            \hline
\multicolumn{1}{|c|}{\textbf{GL-SigRep} \cite{dong2016learning}}   & 0.472 & 12.6\% \\
\multicolumn{1}{|c|}{\textbf{GL-informed}}  & 0.375 & 13.2\% \\
\multicolumn{1}{|c|}{\textbf{GL-conv}}      & 1.240 & 14.8\% \\
\multicolumn{1}{|c|}{\textbf{ML}}           & \textbf{0.347} & \textbf{10.7}\% \\
\hline
\end{tabular}
\end{table}
During this experiment, we set $\gamma = 1000$ for \textbf{ML} and we normalize the volume of the graph obtained by \textbf{GL-conv} to $N$ to provide a fair comparison. Based on the results, \textbf{GL-conv} performs poorly compared to other methods, which can be explained by its lack of adaptability to the given signal set. Recall that it finds a convex combination of the given graph layers in order to fit the smooth signals, which is not very flexible due to the tight search space. \textbf{GL-SigRep}, on the other hand, manages to outperform it by learning the structure directly from the signals. \textbf{GL-Informed} performs better than \textbf{GL-SigRep} in terms of MSE, which indicates that knowing the multi-layer graph representation brings certain advantages. By taking this advantage and coupling it with the flexibility in adapting to the signal set, \textbf{ML} leads to a better inpainting performance than the competitors both in terms of MSE and MAPE.
\subsection{Learning from Social Network Data}
\begin{figure}[!h]
  \centering
  \captionsetup{justification=centering}
  \includegraphics[width=0.5\textwidth]{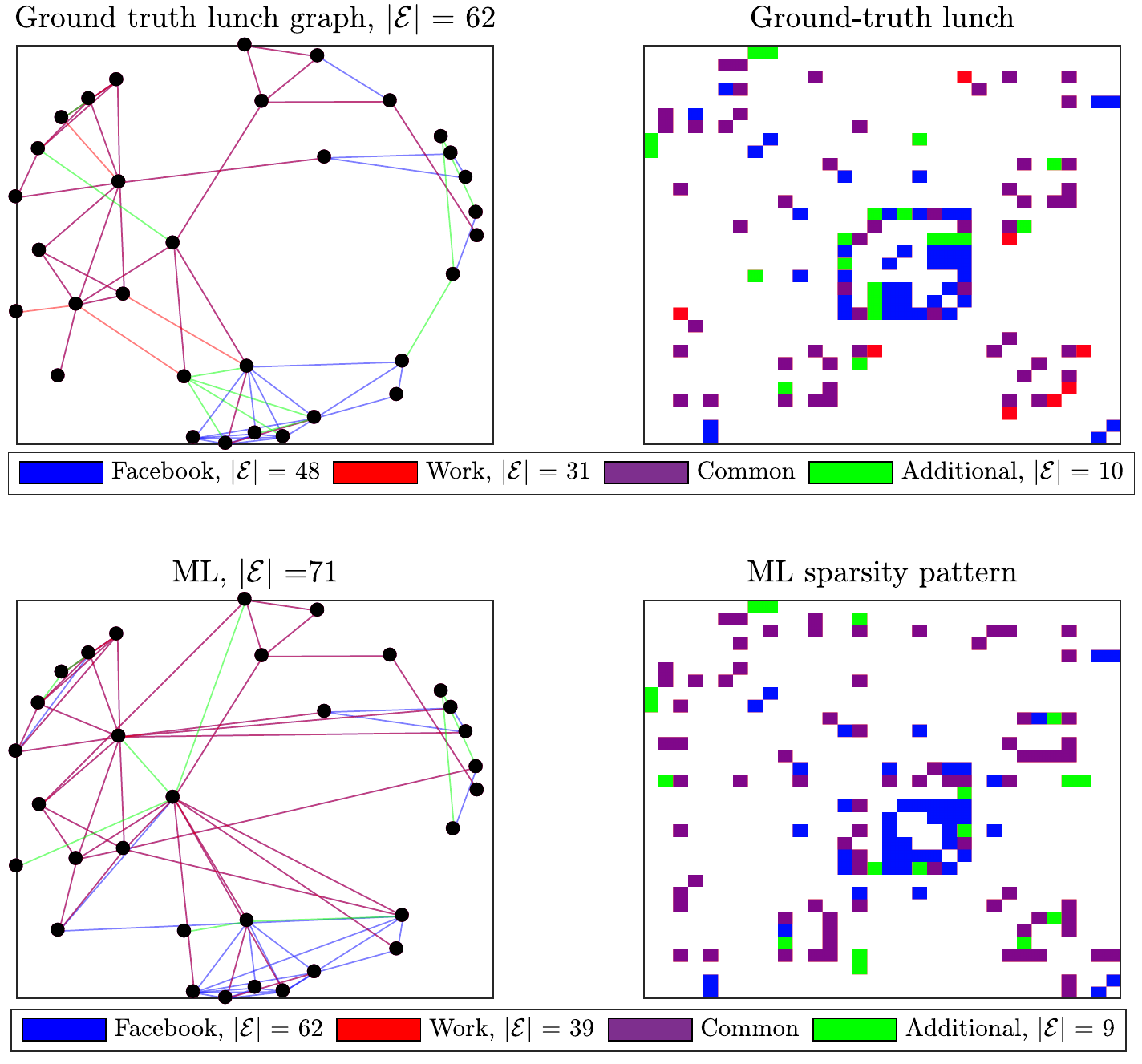}
	\caption{Performance of \textbf{ML} ($\gamma = 0.6, \Gamma = 32$) on CS-AARHUS data}
  \label{fig:cs_ml}
\end{figure}

\begin{figure}[!h]
  \centering
  \captionsetup{justification=centering}
  \includegraphics[width=0.35\textwidth]{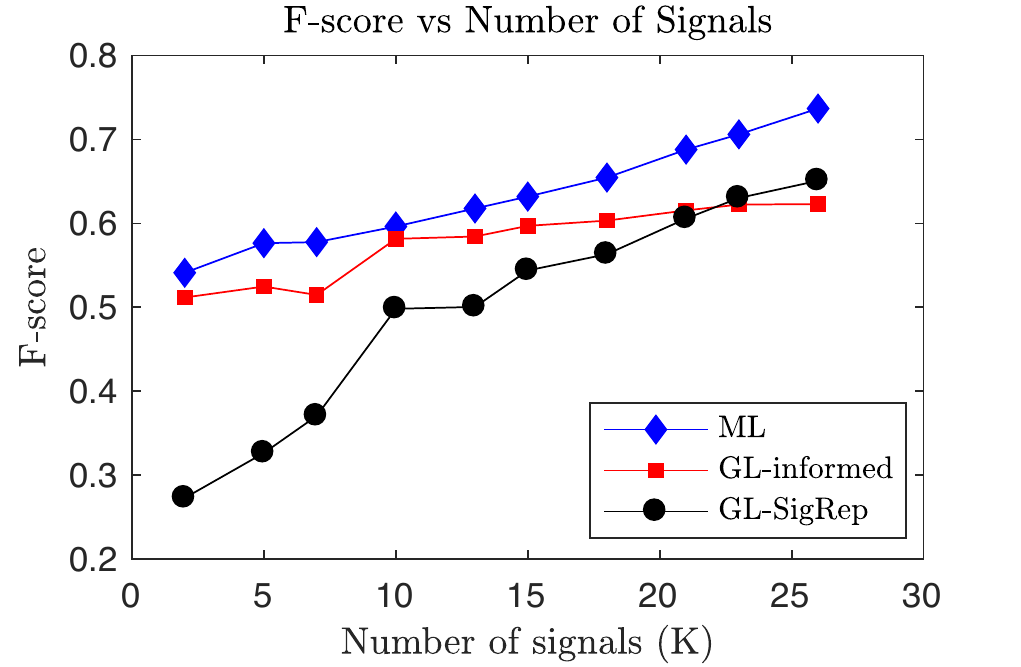}
	\caption{Performance of the graph learning algorithms vs number of signals in lunch data}
  \label{fig:numsig_cs}
\end{figure}

Finally, we test our algorithm on the social network dataset\footnote{http://deim.urv.cat/~alephsys/data.html} provided by \cite{magnani2013combinatorial}. It consists of five kinds of relationship data among 62 employees of the Computer Science Department at Aarhus University (CS-AARHUS), including Facebook, leisure, work, co-authorship and lunch connections. For the experiment, we separate the people into two groups; the first group $\mathcal{A}$ is composed of 32 people having a Facebook account, hence it forms the Facebook network. The second group $\mathcal{B}$ contains any other person eating lunch with anyone in $\mathcal{A}$. The cardinality of $\mathcal{B}$ is 26. We consider a binary matrix $\mathbf{X} \in \mathbb{R}^{32 \times 26}$ that stores the lunch records between groups $\mathcal{A}$ and $\mathcal{B}$ as the signal matrix. Our target task is a graph learning problem where we want to discover the lunch connections inside $\mathcal{A}$ by looking at the lunch records between $\mathcal{A}$ and $\mathcal{B}$. For the graph learning problem, we revive the ``Friend of my friend is my friend." logic through the smoothness of the signal set. In other words, we assume that two people in $\mathcal{A}$ having lunch with the same person in $\mathcal{B}$ will probably have lunch together. Then, via the mask learning scheme, we exploit the Facebook and work connections among people in $\mathcal{A}$. Hence, the inputs of the mask learning algorithm are (i) the multi-layer graph representation formed by the Facebook and work layers composing $\mathcal{A}$, which makes the number of vertices in the graph representation $N=32$, and (ii) the signal set that consists of the lunch records taken on $\mathcal{B}$, which makes the number of signals $K=26$. Then, the output is the lunch network of $\mathcal{A}$. The number of edges is 124 in Facebook layer and 68 in work layer. The coverability of the union of Facebook and work layers on the ground truth lunch network is $0.84$ since the lunch network has 10 connections that do not exist in any of the layers.
The ground truth lunch network and the one inferred by \textbf{ML} are presented in Fig. \ref{fig:cs_ml} together with a color code for the layers.
We compare the performance in terms of the retrieval of the lunch network for the following graph learning algorithms: \textbf{ML}, \textbf{GL-informed}, \textbf{GL-SigRep} and the power-sociomatrix that is introduced by \cite{magnani2013combinatorial}. The performance metrics given in Table \ref{tbl: cs} are calculated with respect to the ground truth lunch network and they measure only the link prediction performance since the networks are unweighted. In addition to the precision, recall and F-score, we use the Jaccard index in order to measure a type of similarity between the inferred graph and the ground truth graph. In \cite{magnani2013combinatorial}, the Jaccard index is computed for two networks to be compared by the proportion of their intersection to their union and it is 1 when the two have identical topology. 
\begin{table}[h]
\caption{Performance of the methods in recovering the lunch network}
\label{tbl: cs}
\resizebox{\columnwidth}{!}{
\begin{tabular}{cc||cccc|}
\cline{3-6}
                   &         & \textbf{Jaccard} &\textbf{ Recall} & \textbf{Precision} & \textbf{F-score}   \\ \hline
\multicolumn{1}{|c|}{\multirow{3}{*}{\parbox{1.3cm}{\textbf{power \newline sociomatrix} \cite{magnani2013combinatorial}}}} & \{FB\}      & 35\%    & 77\%   & 39\%      & 51\% \\
\multicolumn{1}{|c|}{}             & \{Work\}    & 31\%    & 50\%   & 46\%      & 48\% \\
\multicolumn{1}{|c|}{}    & \{FB,Work\} & 34\%    & 84\%   & 37\%      & 51\% \\
\hline
\multicolumn{2}{|c||}{\textbf{GL-SigRep} \cite{dong2016learning}}  &  48\%    & 64\%   & 66\%      & 65\% \\
\hline
\multicolumn{2}{|c||}{\textbf{GL-Informed}}               & 45\%         & 63\%   & 61\%      & 62\% \\
\hline
\multicolumn{2}{|c||}{\textbf{ML}}    &  \textbf{58}\%       & 69\%   & 79\%      & \textbf{74}\% \\
\hline
\end{tabular}
}
\end{table}
Regarding the Jaccard index and the F-score, \textbf{ML} performs best at the recovery of the lunch network by exploiting the multi-layer representation and the signal set at the same time.
With the power-sociomatrix, we obtain all possible combinations of the layers: (i) only the Facebook layer, which is referred to as \{FB\}, (ii) only the work layer, which is referred to as \{Work\}, and (iii) the union of the two layers, which is referred to as \{FB, Work\}. Note that the recall value stated for \{FB, Work\} also gives the coverability of the multi-layer graph representation, which is computed by dividing the number of lunch connections given by the Facebook or the work layer by the total number of lunch connections.
The power-sociomatrix can achieve a limited F-score and Jaccard index since it depends on a simple merging of the two layers without an edge selection process. Then, despite the reasonable coverability rate, \textbf{GL-informed} can not reach the performance of \textbf{GL-SigRep}, which implies that the signal representation quality is better than the multi-layer representation quality to reach the global graph structure. Yet, when we repeat the experiment with signal sets with a lower number of signals, we observe that \textbf{GL-informed} outperforms \textbf{GL-SigRep} when the multi-layer graph representation becomes more informative than the signals. The related results are plotted in Fig. \ref{fig:numsig_cs}, where we train the algorithms with different numbers of signals, $K$, at each experiment. Here, the signal set is randomly formed from the lunch records on $\mathcal{B}$ with the corresponding $K$, and the F-score is averaged over 10 such instances. We employ \textbf{ML} in reduced version \eqref{eqn:ML-LP} when $K < 10$ so that it depends more on the multi-layer graph representation to compensate for the lack of knowledge from the signal side. This permits \textbf{ML} to have the adaptability to different conditions and to outperform the competitor methods as seen in Fig. \ref{fig:numsig_cs}.

\section{Conclusion}
\label{sec: Conclusion}
In this paper, we have introduced a novel method to learn a global graph that explains the structure of a set of smooth signals using partial information provided by multi-layer graphs.
In comparison to the state-of-the-art graph learning methods, the proposed algorithm accepts an additional input, which is the multi-layer graph representation of the data space. This permits to profit from additional domain knowledge in the learning procedure. Our new graph inference algorithm flexibly adjusts the learning procedure between the signal representation and the multi-layer graph representation model, which permits adapting to the quality of the input data.
The algorithm further outputs the mask combination of the layers, which indicates the relative relevance of the multi-layer graphs in inferring the global structure of the signals.

The future work will focus on different signal representation models that can be defined by a multi-layer graph representation.
Moreover, different masking techniques can also be developed considering some examples encountered in real-world data, such as node-independent or locally consistent masking of the graph layers.

\ifCLASSOPTIONcaptionsoff
  \newpage
\fi



\bibliographystyle{IEEEtran}
\bibliography{IEEEabrv,References}
%
%
%

%





\end{document}